\documentclass[11pt,a4paper]{article}
\usepackage[hyperref]{acl2021}
\usepackage{times}
\usepackage{latexsym}
\usepackage{amsmath}
\usepackage{booktabs}
\usepackage{xspace}
\usepackage{siunitx}
\usepackage{graphicx}

\newcommand{\into}{$\rightarrow$}

\newcommand{\ours}{\textsc{LangVarMT}\xspace}

\usepackage[T2A,T1]{fontenc}
\usepackage[utf8]{inputenc}
\usepackage[russian,arabic,english]{babel}
\usepackage{microtype}

\aclfinalcopy 
\def\aclpaperid{3415} 

\newcommand{\Sref}[1]{\S\ref{#1}}

\newcommand{\Tref}[1]{Table~\ref{#1}}
\newcommand{\Aref}[1]{Appendix~\ref{#1}}

\usepackage{xcolor}

\title{Machine Translation into Low-resource Language Varieties}

\author{Sachin Kumar$^\clubsuit$ \quad Antonios Anastasopoulos$^\diamondsuit$ \quad Shuly Wintner$^\heartsuit$ \quad Yulia Tsvetkov$^\spadesuit$ \\
$^\clubsuit$Language Technologies Institute, Carnegie Mellon University, Pittsburgh, PA, USA \\
$^\diamondsuit$Department of Computer Science, George Mason University, Fairfax, VA, USA \\
$^\heartsuit$Department of Computer Science, University of Haifa, Haifa, Israel \\
$^\spadesuit$Paul G.~Allen School of Computer Science \& Engineering, University of Washington \\
\texttt{\small sachink@cs.cmu.edu, antonis@gmu.edu, shuly@cs.haifa.ac.il, yuliats@cs.washington.edu}}

\date{}

\begin{document}
\maketitle

\begin{abstract}
State-of-the-art machine translation (MT) systems are typically trained to generate ``standard'' target language; however, many languages have multiple varieties (regional varieties, dialects, sociolects, non-native varieties) that are different from the standard language. Such varieties are often low-resource, and hence do not benefit from contemporary NLP solutions, MT included.
We propose a general framework to rapidly adapt MT systems to generate language varieties that are close to, but different from, the standard target language, using no parallel (source--variety) data. This also includes adaptation of MT systems to low-resource typologically-related target languages.\footnote{Code, data and trained models are available here: \url{https://github.com/Sachin19/seq2seq-con}}
%
We experiment with adapting an English--Russian MT system to generate Ukrainian and Belarusian, an English--Norwegian Bokmål system to generate Nynorsk, and an English--Arabic system to generate four Arabic dialects, obtaining significant improvements over competitive baselines.
\end{abstract}

\section{Introduction}

Despite tremendous progress in machine translation~\citep{nmtattend,46201} and language generation in general, current 
state-of-the-art 
systems often work under the assumption that a language is homogeneously spoken and understood by its speakers: they generate a ``standard'' form of the target language, typically based on the availability of parallel data. But language use varies with regions, socio-economic backgrounds, ethnicity, and fluency, and many widely spoken languages consist of dozens of varieties or dialects,
with differing lexical, morphological, and syntactic patterns for which no translation data are typically available. As a result,  models trained to translate from a source language (\textsc{src}) to a standard language variety (\textsc{std}) lead to a sub-par experience for speakers of other varieties.

Motivated by these issues, we focus on the task of adapting a trained \textsc{src\into std} translation model to generate text in a different target variety (\textsc{tgt}), having access only to limited monolingual corpora in \textsc{tgt} and no \textsc{src--tgt} parallel data. \textsc{tgt} may be a dialect of, a language variety of, or a typologically-related language to \textsc{std}.

We present an effective transfer-learning framework for translation into low resource language varieties. Our method reuses \textsc{src\into std} MT models and finetunes them on 
synthesized (pseudo-parallel)
\textsc{src--tgt} texts. This allows for rapid adaptation of MT models to new varieties without having to train everything from scratch. Using word-embedding adaptation techniques, we show that MT models which predict continuous word vectors~\citep{kumar2018vmf} rather than softmax probabilities lead to superior performance since they allow additional knowledge to be injected into the models through transfer between word embeddings of high-resource (\textsc{std}) and low-resource (\textsc{tgt}) monolingual corpora.

We evaluate our framework on three translation tasks: English to Ukrainian and Belarusian, assuming parallel data are only available for English\into Russian; English to Nynorsk, with only English to Norwegian Bokmål parallel data; and English to four Arabic dialects, with only English\into Modern Standard Arabic (MSA) parallel data. Our approach outperforms competitive baselines based on unsupervised MT, and methods based on finetuning softmax-based models.


\begin{figure*}
    \centering
    \includegraphics[width=0.8\textwidth]{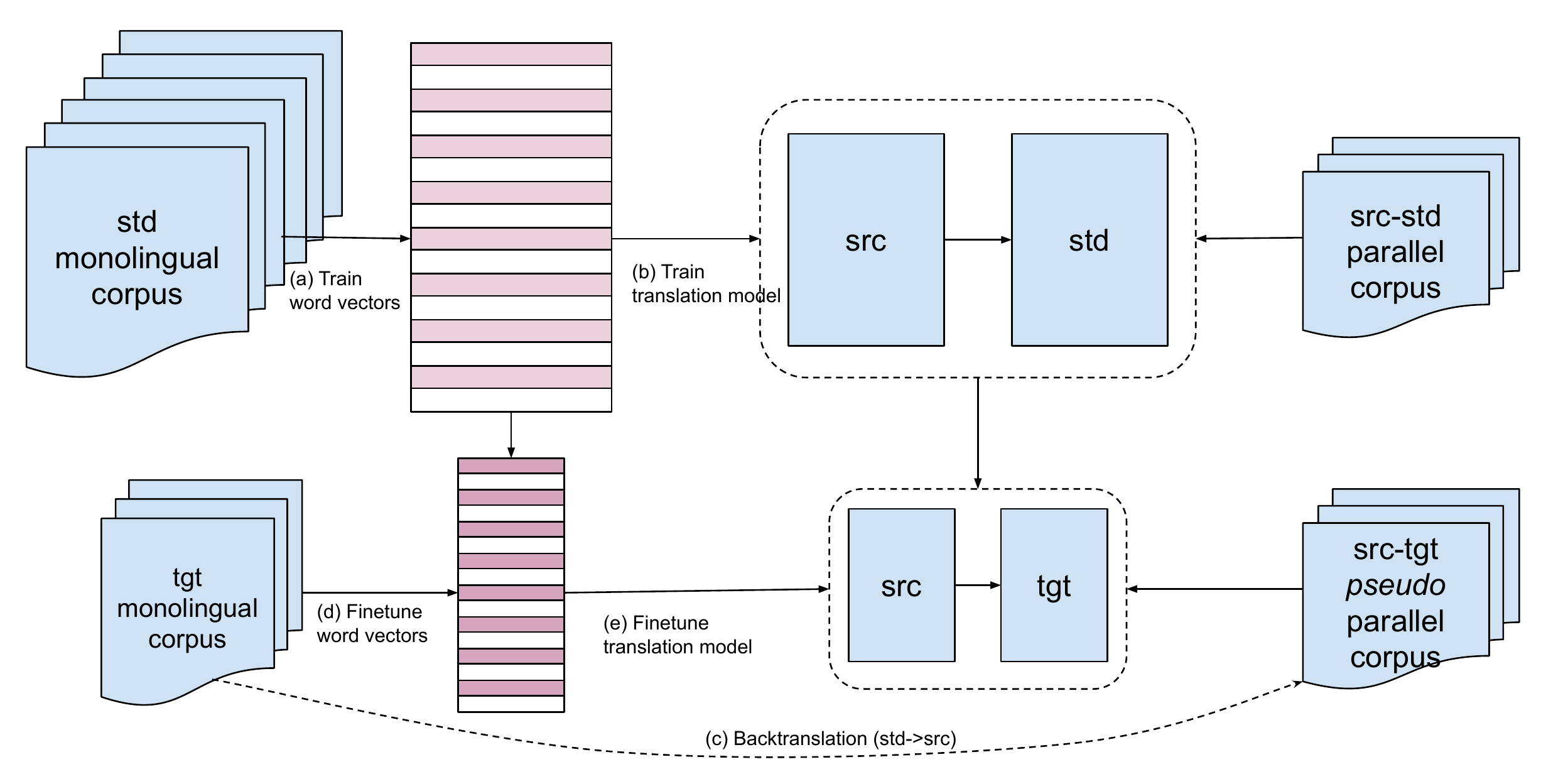}
    \caption{\small An overview of our approach. (a) Using the available \textsc{std} monolingual corpora, we first train word vectors using \texttt{fasttext}; (b) we then train a \textsc{src\into std} translation model using the parallel corpora to predict the pretrained word vectors; (c) next, we train \textsc{std\into src} model and use it to translate \textsc{tgt} monolingual corpora to \textsc{src}; (d) now, we finetune \textsc{std} subword embeddings to learn \textsc{tgt} word embeddings; and finally (e) we finetune a \textsc{src\into std} model to generate \textsc{tgt} pretrained embeddings using the back-translated \textsc{src\into tgt} data.}
    \label{fig:overview}
    \vspace{-1em}
\end{figure*}
\section{A Transfer-learning Architecture}
\label{sec:framework}


We first formalize the task setup. We are given 
a parallel \textsc{src\into std} corpus,
which allows us to train a translation model $f(\cdot; \theta)$ that takes an input sentence $x$ in \textsc{src} and generates its translation in the standard veriety \textsc{std}, $\hat{y}_\textsc{std} = f(x; \theta)$. Here, $\theta$ are the learnable parameters of the model. 
We are also given monolingual corpora in both the standard \textsc{std}
and target variety \textsc{tgt}. 
Our goal now is to
modify $f$ to generate translations $\hat{y}_\textsc{tgt}$ in the target variety \textsc{tgt}. At training time, we assume no \textsc{src}--\textsc{tgt} or \textsc{std}--\textsc{tgt} parallel data are available.

Our solution  (Figure~\ref{fig:overview}) is based on a transformer-based encoder-decoder architecture~\citep{46201} which we modify to predict word vectors. 
Following \citet{kumar2018vmf}, instead of treating each token in the vocabulary as a discrete unit, we represent it using a unit-normalized $d$-dimensional pre-trained vector. These vectors are learned from a \textsc{std} monolingual corpus using \texttt{fasttext}~\citep{bojanowski2016enriching}. A word's representation is computed as the average of the vectors of its character $n$-grams, allowing surface-level linguistic information to be shared among words. 
%
At each step in the decoder, we feed this pretrained vector at the input and instead of predicting a probability distribution over the vocabulary using a softmax layer, we predict a $d$-dimensional continuous-valued vector.
We train this model by minimizing the von Mises-Fisher (vMF) loss---a probabilistic variant of cosine distance---between the predicted vector and the pre-trained vector. The pre-trained vectors (at both input and output of the decoder) are not trained with the model.
To decode from this model, at each step, the output word is generated by finding the closest neighbor (in terms of cosine similarity) of the predicted output vector in the pre-trained embedding table.

We train $f$ in this fashion using \textsc{src--std} parallel data. As shown below, training a softmax-based $\textsc{src\into std}$ model to later finetune with \textsc{tgt} suffers from vocabulary mismatch between \textsc{std} and \textsc{tgt} and thus is detrimental to downstream performance. By replacing the decoder input and output with pretrained vectors, we separate the vocabulary from the MT model, making adaptation easier.

Now, to finetune this model to generate \textsc{tgt}, we need \textsc{tgt} embeddings. Since the \textsc{tgt} monolingual corpus is small, training \texttt{fasttext} vectors on this corpus from scratch will lead (as we show) to low-quality embeddings. Leveraging the relatedness of \textsc{std} and \textsc{tgt} and their vocabulary overlap, we use \textsc{std} embeddings to transfer knowledge to \textsc{tgt} embeddings: for each character $n$-gram in the \textsc{tgt} corpus, we initialize its embedding with the corresponding \textsc{std} embedding, if available. We then continue training \texttt{fasttext} on the \textsc{tgt} monolingual corpus~\citep{aditi}. Last, we use a supervised embedding alignment method~\citep{lample2017unsupervised} to project the learned \textsc{tgt} embeddings in the same space as \textsc{std}. \textsc{std} and \textsc{tgt} are expected to have a large lexical overlap, so we use identical tokens in both varieties as supervision for this alignment. The obtained embeddings, 
due to transfer learning from \textsc{std}, inject additional knowledge in the model. 

Finally, to obtain a \textsc{src\into tgt} model, we finetune $f$ on psuedo-parallel \textsc{src--tgt} data. Using a \textsc{std\into src} MT model (a back-translation model trained using large \textsc{std--src} parallel data with standard settings) we (back)-translate \textsc{tgt} data to \textsc{src}.
Naturally, these synthetic parallel data will be noisy despite the similarity between \textsc{std} and \textsc{tgt}, but we show that they improve the overall performance. We discuss the implications of this noise in \Sref{sec:results}. 
\section{Experimental Setup}
\label{sec:method}


\paragraph{Datasets}
We experiment with two setups. In the first (synthetic) setup, we use English (\textsc{en}) as \textsc{src}, Russian (\textsc{ru}) as \textsc{std}, and Ukrainian (\textsc{uk}) and Belarusian (\textsc{be}) as \textsc{tgt}s. We sample 10M \textsc{en-ru} sentences from the WMT'19 shared task~\citep{ma-etal-2019-results}, and 80M \textsc{ru} sentences from the \href{http://universaldependencies.org/conll17/data.html}{CoNLL'17 shared task} to train embeddings. To simulate low-resource scenarios, we sample 10K, 100K and 1M \textsc{uk} sentences from the CoNLL'17 shared task and \textsc{be} sentences from the \href{https://oscar-corpus.com/}{OSCAR corpus}~\citep{ortiz-suarez-etal-2020-monolingual}. We use TED dev/test sets for both languages pairs~\citep{cettolo-etal-2012-wit3}. 

The second (real world) setup has two language sets: the first one defines English as \textsc{src}, with Modern Standard Arabic (\textsc{msa}) as \textsc{std} and four Arabic varieties spoken in {Doha}, {Beirut}, {Rabat} and {Tunis} as \textsc{tgt}s. We sample 10M \textsc{en-msa} sentences from the UNPC corpus~\citep{ziemski-etal-2016-united}, and 80M \textsc{msa} sentences from the CoNLL'17 shared task. For Arabic varieties, we use the MADAR corpus~\citep{bouamor-etal-2018-madar} which consists of 12K 6-way parallel sentences between English, MSA and the 4 considered varieties. We ignore the English sentences, sample dev/test sets of 1K sentences each,  and consider 10K monolingual sentences for each \textsc{tgt} variety. The second set also has English as \textsc{src} with Norwegian Bokmål (\textsc{no}) as \textsc{std} and its written variety Nynorsk (\textsc{nn}) as \textsc{tgt}. We use 630K \textsc{en-no} sentences from WikiMatrix~\citep{schwenk2019wikimatrix}, and 26M \textsc{no} sentences from ParaCrawl~\citep{espla-etal-2019-paracrawl} combined with the WikiMatrix \textsc{no} sentences to train embeddings. We use 310K \textsc{nn} sentences from WikiMatrix, and TED dev/test sets for both varieties~\citep{reimers-2020-multilingual-sentence-bert}. 


\paragraph{Preprocessing} We preprocess raw text using Byte Pair Encoding \citep[\href{https://github.com/glample/fastBPE}{BPE},][]{sennrich-etal-2016-neural}
with 24K merge operations on each \textsc{src--std} corpus trained separately on \textsc{src} and \textsc{std}. 
We use the same BPE model to tokenize the monolingual \textsc{std} data and learn \texttt{fasttext} embeddings (we consider character $n$-grams of length 3 to 6).\footnote{We slightly modify \texttt{fasttext} to not consider BPE token markers ``@@'' in the character $n$-grams.} 
Splitting the \textsc{tgt} words with the same \textsc{std} BPE model will result in heavy segmentation, especially when \textsc{tgt} contains characters not present in \textsc{std}.\footnote{For example, both \textsc{ru} and \textsc{uk} alphabets consist of 33 letters; \textsc{ru} has the letters {\selectlanguage{russian} Ёё, ъ, ы and Ээ}, which are not used in \textsc{uk}. Instead, \textsc{uk} has {\selectlanguage{russian} Ґґ, Єє, Іі and Її}.} 
To counter this, we train a joint BPE model with 24K operations on the concatenation of \textsc{std} and \textsc{tgt} corpora to tokenize \textsc{tgt} corpus following \citet{chronopoulou-etal-2020-reusing}. 
This technique increases the number of shared tokens between \textsc{std} and \textsc{tgt}, thus enabling better cross-variety transfer while learning embeddings \emph{and} while finetuning. We follow~\citet{aditi} to train embeddings on the generated \textsc{tgt} vocabulary where we initialize the character $n$-gram representations for \textsc{tgt} words with \textsc{std}'s  \texttt{fasttext} model wherever available and finetune them on the \textsc{tgt} corpus.

\paragraph{Implementation and Evaluation} 
We modify the standard \href{https://github.com/OpenNMT/OpenNMT-py}{OpenNMT-py} seq2seq models of PyTorch~\citep{klein-etal-2017-opennmt} to train our model with vMF loss~\citep{kumar2018vmf}. 
Additional hyperparameter details are outlined in Appendix~\ref{app:implementation}.
We evaluate our methods using BLEU score~\citep{papineni-etal-2002-bleu} based on the \href{https://github.com/mjpost/sacreBLEU}{SacreBLEU} implementation~\citep{post-2018-call}.\footnote{While we recognize the limitations of BLEU~\citep{mathur-etal-2020-tangled}, more sophisticated embedding-based metrics for MT evaluation~\citep{Zhang*2020BERTScore:,sellam-etal-2020-bleurt} are unfortunately not available for low-resource language varieties.} For the Arabic varieties, we also report a macro-average. In addition, to measure the expected impact on actual systems' users, we follow~\citet{faisal-etal-21-sdqa} in computing a population-weighted macro-average (\textbf{$\text{avg}_{\text{pop}}$}) based on language community populations provided by Ethnologue~\cite{eberhard22simons}.

\subsection{Experiments}

\begin{table*}[t]
\small
\centering
\begin{tabular}{lSSS|SSS|S|SSSS}
\toprule
& \multicolumn{3}{c|}{\textbf{\textsc{uk}}} & \multicolumn{3}{c|}{\textbf{\textsc{be}}} & \multicolumn{1}{c|}{\textbf{\textsc{nn}}} &  \multicolumn{4}{c}{\textbf{Arabic Varieties} (10K)} \\
Size of \textsc{tgt} corpus& \multicolumn{1}{c}{10K} & \multicolumn{1}{c}{100K} & \multicolumn{1}{c|}{1M} & \multicolumn{1}{c}{10K} & \multicolumn{1}{c}{100K} & \multicolumn{1}{c|}{1M} & \multicolumn{1}{c|}{300K}  & \textbf{Doha} & \textbf{Beirut} & \textbf{Rabat} & \textbf{Tunis}\\ \midrule
\textsc{Sup(src\into std)} & 1.7 & 1.7 & 1.7 & 1.5 & 1.5 & 1.5 & 11.3   & 3.7  & 1.8 & 2.0 & 1.3 \\
\textsc{Unsup(src\into tgt)} & 0.3	& 0.6 & 0.9 & 0.4 & 0.6 & 1.4 & 2.7 & 0.2 & 0.1 & 0.1 & 0.1  \\
\textsc{Pivot} & 1.5 & 8.6 & 14.9 & 1.15 & 3.9 & 8.0 & 11.9 & 1.8 & 2.1 & 1.7 & 1.1\\
\textsc{Softmax} & 1.9 & 12.7 & 15.4 & 1.5 &  4.5 & 7.9 & 14.4  & 14.5 & 7.4 & 4.9 & 3.9 \\
\textbf{\ours}  & {\textbf{6.1}} & {\textbf{13.5}} & {\textbf{15.3}} & {\textbf{2.3}} & {\textbf{8.8}} & {\textbf{9.8}} & \textbf{16.6} & {\textbf{20.1}} & {\textbf{8.1}} & {\textbf{7.4}} & {\textbf{4.6}} \\ 
\bottomrule
\end{tabular}
\caption{\small BLEU scores on translation from English to Ukrainian, Belarusian, Nynorsk, and Arabic dialects with varying amounts of monolingual target data (\textsc{tgt} sentences)  available for finetuning. Our approach (\ours) outperforms all baselines.}
\label{tab:data-size-comparison}
\vspace{-1em}
\end{table*}

Our proposed framework, \textbf{\ours}, consists of three main components:
(1) A supervised \textsc{src\into std} model is trained to predict continuous \textsc{std} word embeddings rather than discrete softmax probabilities. 
(2) Output \textsc{std} embeddings are replaced with \textsc{tgt} embeddings. The \textsc{tgt} embeddings are trained by finetuning \textsc{std} embeddings on monolingual \textsc{tgt} data and aligning the two embedding spaces.
(3) The resulting model is finetuned with pseudo-parallel \textsc{src\into tgt} data.

We compare \ours with the following competitive baselines. 
\textbf{\textsc{Sup(src\into std)}}: train a standard (softmax-based) supervised \textsc{src\into std} model, and consider the output of this model as \textsc{tgt} under the assumption that \textsc{std} and \textsc{tgt} may be very similar.
\textbf{\textsc{Unsup(src\into tgt)}}:  train an unsupervised MT model~\citep{lample2017unsupervised} in which the encoder and decoder are initialized with cross-lingual masked language models \citep[MLM,][]{NEURIPS2019_c04c19c2}. These MLMs are pre-trained on \textsc{src} monolingual data, and then finetuned on \textsc{tgt} monolingual data with an expanded vocabulary as described above. This baseline is taken from~\citet{chronopoulou-etal-2020-reusing}, where it showed state-of-the-art performance for low-monolingual-resource scenarios.
\textbf{Pivot}: train a \textsc{Unsup({std\into tgt})} model as described above using \textsc{std} and \textsc{tgt} monolingual corpora. During inference, translate the \textsc{src} sentence to \textsc{std} with the \textsc{Sup(src\into std)} model and then to \textsc{tgt} with the \textsc{Unsup({std\into tgt})} model. 
We also perform several ablation experiments, showing that every component of \ours is necessary for good downstream performance. Specifically, we report results with \ours but using a standard softmax layer (\textbf{\textsc{softmax}}) to predict tokens instead of continuous vectors.
\footnote{Additional ablation results are listed in Appendix~\ref{app:ablations}.}

\section{Results and Analysis}
\label{sec:results}

\Tref{tab:data-size-comparison} compares the performance of \ours\ with the baselines for Ukrainian, Belarusian, Nynorsk, and the four Arabic varieties. For reference, note that the \textsc{en\into ru}, \textsc{en\into msa}, and \textsc{en\into no} models are relatively strong, yielding BLEU scores of $24.3$, $21.2$, and $24.9$, respectively.

\paragraph{Synthetic Setup}
Considering \textsc{std} and \textsc{tgt} as the same language is sub-optimal, as is evident from the poor performance of the non-adapted \textsc{Sup(src\into std)} model. Clearly, special attention ought to be paid to language varieties. 
Direct unsupervised translation from \textsc{src} to \textsc{tgt} performs poorly as well, confirming previously reported results of the ineffectiveness of such methods on unrelated languages~\cite{guzman-etal-2019-flores}.

Translating \textsc{src} to \textsc{tgt} by pivoting through \textsc{std} achieves much better performance owing to strong \textsc{Unsup(std\into tgt)} models that leverage the similarities between \textsc{std} and \textsc{tgt}. However, when resources are scarse (e.g., with 10K monolingual sentences as opposed to 1M), this performance gain considerably diminishes. We attribute this drop to overfitting during the pre-training phase on the small \textsc{tgt} monolingual data. Ablation results  (Appendix~\ref{app:ablations}) also show that in such low-resource settings the learned embeddings are of low quality.

Finally, \ours consistently outperforms all baselines. Using 1M \textsc{uk} sentences, it achieves similar performance (for \textsc{en\into uk}) to the softmax ablation of our method, \textsc{Softmax}, and small gains over unsupervised methods. However, in lower resource settings our approach is clearly better than the strongest baselines by over 4 BLEU points for \textsc{uk} (10K) and 3.9 points for \textsc{be} (100K). 

To identify potential sources of error in our proposed method, we lemmatize the generated translations and test sets and evaluate BLEU~\citep{qi2020stanza}. Across all data sizes, both \textsc{uk} and \textsc{be} achieve a substantial increase in BLEU (up to +6 BLEU; see \Aref{app:analysis} for details) compared to that obtained on raw text, indicating morphological errors in the translations. In future work, we will investigate whether we can alleviate this issue by considering \textsc{tgt} embeddings based on morphological features of
tokens~\citep{aditi}.

\paragraph{Real-world Setup} 
The effectiveness of \ours is pronounced in this setup with a dramatic improvement of more than 18 BLEU points over unsupervised baselines when translating into Doha Arabic. We hypothesize that during the pretraining phase of unsupervised methods, the extreme difference between the size of the \textsc{msa} monolingual corpus (10M) and the varieties' corpora (10K) leads to overfitting. Additionally, compared to the synthetic setup, the Arabic varieties we consider are quite close to \textsc{msa}, allowing for easy and effective adaptation of both word embeddings and \textsc{en\into msa} models. 
\ours\ also improves in all other Arabic varieties, although naturally some varieties remain challenging. For example, the Rabat and particularly the Tunis varieties are more likely to include French loanwords~\cite{bouamor-etal-2018-madar} which are not adequately handled as they are not part of our vocabulary. 
In future work, we will investigate whether we can alleviate this issue by potentially including French corpora (transliterated into Arabic) to our \textsc{tgt} language corpora. 
On average, our approach improves by 2.3 BLEU points over the softmax-based baseline (cf. 7.7 and 10.0 in Table~\ref{tab:fairness} under \textbf{$\text{avg}_{\mathcal{L}}$}) across the four Arabic dialects. 
For a population-weighted average (\textbf{$\text{avg}_{\text{pop}}$}), we associate the Doha variety with Gulf Arabic (ISO code: \texttt{afb}), the Beirut one with North Levantine Arabic (\texttt{apc}), Rabat with Moroccan  (\texttt{ary}), and the Tunis variety with Tunisian Arabic  (\texttt{aeb}). As before, \ours{} outperforms the baselines.
The absolute BLEU scores in this highly challenging setup are admittedly low, but as we discuss in \Aref{app:analysis}, the translations generated by \ours\ are often fluent and input preserving, especially compared to the baselines.

Finally, due to high similarity between \textsc{no} and \textsc{nn}, the \textsc{Sup(en\into no)} model also performs well on \textsc{nn} with 11.3 BLEU, but our method yields further gains of over 4 points over the baselines.





\section{Discussion}
\noindent
\textbf{Fairness}
The goal of this work is to develop more equitable technologies, usable by speakers of diverse language varieties. Here, we evaluate the systems along the principles of \emph{fairness}. 
We evaluate the fairness of our Arabic multi-dialect system's utility proportionally to the populations speaking those dialects. In particular, we seek to measure how much average benefit will the people of different dialects receive if their respective translation performance is improved.
A simple proxy for fairness is the standard deviation (or, even simpler, a $\max-\min$ performance) of the BLEU scores across dialects (A higher value implies more unfairness across the dialects) Beyond that, we measure a system's \textit{unfairness} with respect to the different dialect subgroups, using the adaptation of generalized entropy index \cite{speicher2018unified}, 
which considers equities within and between subgroups in evaluating the overall unfairness of an algorithm on a population \citet{faisal-etal-21-sdqa} (See Appendix~\ref{app:unfairness} for details and additional discussion).
Table~\ref{tab:fairness} shows that our proposed method is fairer 
across all dialects, compared to baselines where only \textsc{msa} translation produces comprehensible outputs.

\begin{table}[htb]
\scriptsize
\centering
\begin{tabular}{@{}l|cc|c@{ }c@{}}
\toprule
Model & \textbf{$\text{avg}_{\mathcal{L}}$}$\uparrow$ & \textbf{$\text{avg}_{\text{pop}}$}$\uparrow$ & \textbf{max$-$min}$\downarrow$ & \textbf{unfair}$\downarrow$\\ \midrule
\textsc{Sup(src\into std)} & 2.2 & 1.8 & 19.9 & 0.037 \\
\textsc{Unsup(src\into tgt)} & 0.1 & 0.1 & 21.1 & 0.046\\
\textsc{Pivot} & 1.7 & 1.8 & 20.1 & 0.037\\
\textsc{Softmax} &  7.7 & 5.7 & 17.3 & 0.020 \\
\textbf{\ours} & \textbf{10.0} & \textbf{7.3} & \textbf{16.6} & \textbf{0.016} \\ 
\bottomrule
\end{tabular}
\caption{\small Average performance and fairness metrics across the four Arabic varieties. This evaluation includes \textsc{msa} (with a BLEU score of 21.2 on the \textsc{sup(en\into msa)} model).}
\label{tab:fairness}
\vspace{-1em}
\end{table}


\noindent
\textbf{Negative Results}
Our proposed method relies on two components: (1) quality of \textsc{tgt} word embeddings which is dependent on \textsc{std} and \textsc{tgt} shared (subword) vocabulary, and (2) the psuedo-parallel \textsc{src--tgt} obtained by back-translating \textsc{tgt} data through a \textsc{std\into src} model. 
If \textsc{std} and \textsc{tgt} are not sufficiently closely related, the quality of both of these components can degrade, leading to a drop in the performance of our proposed method. We present results of two additional experiments to elucidate this phenomenon in \Aref{app:negative-results}.

\noindent
\textbf{Related Work} We provide an extensive discussion of related work in Appendix~\ref{app:related}.


\section{Conclusion}

We presented a transfer-learning framework for rapid and effective adaptation of MT models to different varieties of the target language without access to any source-to-variety parallel data. We demonstrated significant gains in BLEU scores across several language pairs, especially in highly resource-scarce scenarios. The improvements are mainly due to the benefits of continuous-output models over softmax-based generation. Our analysis highlights the importance of addressing morphological differences between language varieties, which will be in the focus of our future work.

\section*{Acknowledgements}
This research was supported by Grants No.\ 2017699 and~2019785 from the United States-Israel Binational Science Foundation (BSF), by the National Science Foundation (NSF) under Grants No.~2040926 and 2007960, and by a Google faculty research award. 
We thank Safaa Shehadi for evaluating our model outputs, Xinyi Wang and Aditi Choudhary for helpful discussions, and the anonymous reviewers for much appreciated feedback.

\clearpage
\bibliographystyle{acl_natbib}
\bibliography{acl2021}

\clearpage
\appendix

\section{Related Work}
\label{app:related}

Early work addressing translation involving language varieties includes rule-based transformations~\citep{altintasmachine, marujo-etal-2011-bp2ep, 6473708} 
which rely on language specific information and expert knowledge which can be expensive and difficult to scale.
Recent work to address this issue
only focuses on cases where parallel data do exist. They include a combination of word-level and character-level MT~\citep{10.5555/1626355.1626360, tiedemann-2009-character, nakov-tiedemann-2012-combining} between related languages or training multilingual models to translate to/from English to different varieties of a language 
(e.g.,~\citet{lakew-etal-2018-neural} work on Brazilian--European Portuguese and European--Canadian French). Such parallel data, however, are typically unavailable for most language varieties.

Unsupervised translation models, which require only monolingual data, can address this limitation~\citep{artetxe2018unsupervised,lample2017unsupervised, garcia-etal-2020-multilingual, garcia2020harnessing}. However, when even \emph{mono}lingual corpora are limited, unsupervised models are challenging to train and are quite ineffective for translating between unrelated languages~\citep{marchisio2020does}. Considering varieties of a language as writing styles, unsupervised style transfer~\citep{NEURIPS2018_398475c8, He2020A} or deciphering methods~\citep{pourdamghani-knight-2017-deciphering} to translate between different varieties have also been been explored but have not been shown to perform well, often only reporting BLEU-1 scores since they obtain BLEU-4 scores which are closer to 0. Additionally, 
all of these approaches require simultaneous access to data in all varieties during training and must be trained from scratch when a new variety is added. In contrast, our presented method allows for easy adaptation of \textsc{src\into std} models to any new variety as it arrives. 

Considering a new target variety as a new domain of \textsc{std}, unsupervised domain adaptation methods can be employed, such as finetuning \textsc{src\into std} models using pseudo-parallel corpora generated from monolingual corpora in target varieties~\citep{hu-etal-2019-domain,currey-etal-2017-copied}. 
Our proposed method is most related to this approach; but
while these methods have the potential to adapt the decoder language model, for effective transfer, \textsc{std} and \textsc{tgt} must have a shared vocabulary which is not true for most language varieties due to lexical, morphological, and at times orthographic differences. In contrast, our proposed method makes use of cross-variety word embeddings.
While our examples only involve same-script varieties, augmenting our approach to work across scripts through a transliteration component is straightforward.

\section{Implementation Details}
\label{app:implementation}

We modify the standard \href{https://github.com/OpenNMT/OpenNMT-py}{OpenNMT-py} seq2seq models of PyTorch~\citep{klein-etal-2017-opennmt} to train our model with vMF loss~\citep{kumar2018vmf}. 
We use the transformer-\textsc{base} model~\citep{46201}, with 6 layers in both encoder and decoder and with 8 attention heads, as our underlying architecture. We modify this model to predict pretrained \texttt{fasttext} vectors. We also initialize the decoder input embedding table with the pretrained vectors and do not update them during model training.  All models are optimized using Rectified Adam~\citep{Liu2020On} with a batch size of 4K tokens and dropout of $0.1$. We train \textsc{src}$\to$\textsc{std} models for 350K steps with an initial learning rate of $0.0007$ with linear decay. For finetuning, we reduce the learning rate to $0.0001$ and train for up to $100$K steps. We use early stopping in all models based on validation loss computed every 2K steps. We decode all the softmax-based models with a beam size of 5 and all the vMF-based models greedily. 

We evaluate our methods using BLEU score~\citep{papineni-etal-2002-bleu} based on the \href{https://github.com/mjpost/sacreBLEU}{SacreBLEU} implementation~\citep{post-2018-call}. While we recognize the limitations of BLEU~\citep{mathur-etal-2020-tangled}, more sophisticated embedding-based metrics for MT evaluation~\citep{Zhang*2020BERTScore:,sellam-etal-2020-bleurt} are simply not available for language varieties.

\section{Additional English-Ukrainian Experiments}
\label{app:ablations}
On our resource-richest setup of \textsc{en\into uk} translation using 1M \textsc{uk} sentences and \textsc{ru} as \textsc{std}, we compare our method with the following additional baselines. 
\Tref{tab:all-baselines} presents these results. 

\begin{table}[hbt]
\small
\centering
\begin{tabular}{lr}
\toprule
\multicolumn{1}{c}{Method}    & \multicolumn{1}{c}{BLEU (uk)} \\ \midrule
\textsc{sup(src-std)} & 1.7 \\ 
\textsc{Unsup(src\into tgt)} & 0.9 \\ 
\textsc{Pivot:} & 14.9 \\\midrule
\textsc{Lample-Unsup(src\into tgt)} & 0.4 \\
\textsc{Pivot:Lample-Unsup(std\into tgt)} & 9.0 \\
\textsc{Pivot:DictReplace(std\into tgt)} &  2.9 \\ \midrule
\ours & 15.3 \\ 
 \textsc{LangVarMT} w/ poor embeddings &  4.6\\
\textsc{LangVarMT-random} & 13.1 \\
\textsc{Softmax} & 15.4  \\
\textsc{LangVarMT-random-softmax} & 14.1 \\  \midrule
 
\end{tabular}
\caption{\textsc{bleu} scores on \textsc{en-uk} test corpus with 1M \textsc{uk} monolingual corpus.}
\label{tab:all-baselines}
\end{table}




\textbf{\textsc{Lample-Unsup(src\into tgt)}}: This is another unsupervised model, based on~\citet{lample2017unsupervised} which initializes the input and output embedding tables of both encoder and decoder using cross-lingual word embeddings trained on \textsc{src} and \textsc{tgt} monolingual corpora. The model is trained in a similar manner to~\citet{chronopoulou-etal-2020-reusing} (\textsc{Unsup(src\into tgt)}) with iterative backtranslation and autoencoding. 

\textbf{\textsc{Pivot:Lample(std\into tgt)}}: This baseline is similar to the \textsc{Pivot} baseline, where we replace the unsupervised model with that of~\citet{lample2017unsupervised}.

\textbf{\textsc{Pivot:DictReplace(std\into tgt)}}: Here we first translate \textsc{src} to \textsc{std} using \textsc{Sup(src\into std)}, and then modify the \textsc{std} output to get a \textsc{tgt} sentence as follows: We create a \textsc{std--tgt} dictionary using the embedding map suggested by~\citet{conneau2017word}. This dictionary is created on words tokenized with Moses tokenizer~\citep{hoang-koehn-2008-design} rather than BPE tokens. We replace each token in the generated \textsc{std} sentence which is not in the \textsc{tgt} vocabulary using the dictionary (if available). We consider this baseline to measure lexical vs.\  syntactic/phrase level differences between Russian and Ukrainian. 

In addition to baseline comparison, we report the following ablation experiments. 


(1) To measure transfer from \textsc{std} to \textsc{tgt} embeddings, we finetune the \textsc{Sup(src\into std)} model using \textsc{tgt} embeddings trained from scratch (as opposed to initialized with \textsc{std} embeddings).

(2) To measure the impact of initialization during model finetuning, we compare with a randomly initialized model trained in a supervised fashion on the psuedo-parallel \textsc{src--tgt} data.

\paragraph{Baselines}
On the unsupervised models based on  \citet{lample2017unsupervised}, we observe a similar trend as that of~\citet{chronopoulou-etal-2020-reusing}, where the \textsc{Lample-Unsup(src\into tgt)} model performing poorly (0.4) with substantial gains when pivoting through Russian (9.0 BLEU). 

\textsc{Pivot:DictReplace(std\into tgt)} gains some improvement over considering the output of \textsc{Sup(src\into std)} as \textsc{tgt}, probably due to syntactic similarities between Russian and Ukrainian. This result can potentially be further improved with a human-curated \textsc{ru--uk} dictionary, but such resources are typically not available for the low-resource settings we consider in this paper. 


\paragraph{Ablations} 
As shown in \Tref{tab:all-baselines}, training the \textsc{src\into tgt} model on a randomly initialized model (\textsc{LangVar-random}) results in a performance drop, confirming that transfer learning from a \textsc{src\into std} model is beneficial. Similarly, using \textsc{tgt} embeddings trained from scratch (\ours w/ poor embeddings) results in a drastic performance drop, providing evidence for essential transfer from \textsc{std} embeddings. 


\section{Analysis}
\label{app:analysis}
To better understand the performance of our models, we perform additional analyses.

\paragraph{Lemmatized BLEU}
For \textsc{uk} and \textsc{be}, we lemmatize each word in the test sets and the translations and evaluate BLEU scores. The results, depicted in \Tref{tab:lemmatized-bleu}, very likely indicate that our framework often generates correct lemmas, but may fail on the correct inflectional form of the target words. This highlights the importance of considering morphological differences between language varieties. The high BLEU scores also demonstrate that the resulting translations are quite likely understandable, albeit not always grammatical.

\begin{table}[hbt]
\scriptsize
\centering
\begin{tabular}{lSSS|SSS}
\toprule
& \multicolumn{3}{c|}{\textbf{\textsc{en\into uk}}} & \multicolumn{3}{c}{\textbf{\textsc{en\into be}}} \\
 & \multicolumn{1}{c}{10K} & \multicolumn{1}{c}{100K} & \multicolumn{1}{c|}{1M} & \multicolumn{1}{c}{10K} & \multicolumn{1}{c}{100K} & \multicolumn{1}{c}{1M} \\ \toprule
raw & 6.1 & 13.5 & 15.3 & 2.3 & 8.8 & 9.8 \\ 
lemma & 12.8 & 19.5 & 21.3 & 3.5 & 13.7 & 15.8 \\ \bottomrule
\end{tabular}
\caption{\small BLEU scores on raw vs lemmatized text with \ours.}
\label{tab:lemmatized-bleu}
\end{table}

\paragraph{Translation of Rare Words}
On the outputs of the \textsc{en\into uk} model, trained with 100K \textsc{uk} sentences,
we compute the translation accuracy of words based on their frequency in the \textsc{tgt} monolingual corpus for \ours, our best baseline \textsc{Sup(src\into std)}+\textsc{Unsup(src\into tgt)} and the best performing ablation \textsc{Softmax}. These results, shown in \Tref{tab:frequency}, reveal that \ours is more accurate at translating rare words (with frequency less than 10) compared to the baselines.

\begin{table}[t]
\small
  \centering
  \begin{tabular}{c|ccc}
  \toprule
frequency & \textsc{Pivot} & \textsc{Softmax} & \ours \\ \midrule
1 & 0.0429 & 0.1516 & 0.1812 \\
2 & 0.0448 & 0.2292 & 0.2556 \\
3 & 0.0597 & 0.2246 & 0.2076 \\
4 & 0.0692 & 0.2601 & 0.2962 \\
{[}5,10) & 0.0582 & 0.2457 & 0.2722 \\
{[}10,100) & 0.1194 & 0.2881 & 0.2827 \\
{[}100,1000) & 0.2712 & 0.4537 & 0.4449 \\ \midrule
  \end{tabular}
  \caption{\small Translation accuracies of words based on their frequencies on \textsc{en\into uk} with 100K \textsc{uk} sentences.}
  \label{tab:frequency}
\end{table}

\paragraph{Examples}
We provide some examples of \textsc{en-uk} and \textsc{en}-Beirut Arabic translations generated by the three models in Tables~\ref{tab:examples} and \ref{tab:beirut_examples}.
As evaluated by native speakers of the Beirut Arabic, we find that despite a BLEU score of only 8, in a majority of cases our baseline model is able to generate fluent translations of the input, preserving most of the content, whereas the baseline model ignores many of the content words. We also observe that in some cases, despite predicting in the right semantic space of the pretrained embeddings, it fails to predict the right token, resulting in surface form errors (e.g., predicting adjectival forms of verbs). This phenomenon is known and studied in more detail in \citet{kumar2018vmf}.

\begin{table}[hbt]
\small
  \centering
  \begin{tabular}{r|l}  
  \toprule
  Source & And we never think about the hidden \\ & connection \\
Reference &  \selectlanguage{russian}Та ми ніколи не думаємо про  \\ & \selectlanguage{russian}приховані зв'язки \\
\textsc{Pivot} & \selectlanguage{russian}І ми ніколи не дуємо про \\ & \selectlanguage{russian}приховану зв'язку. \\
& (And we never think about a hidden \\ & connection.) \\
\textsc{Softmax} & \selectlanguage{russian}Я ніколи не думав про \\ & \selectlanguage{russian}прихований зв'язок. \\
& (I never thought of a hidden connection.)\\
\ours & \selectlanguage{russian}І ми ніколи не думаємо про \\ & \selectlanguage{russian}прихований зв'язок. \\
& (And we never think about a hidden \\ & connection.) \\ \midrule
Source & And yet, looking at them, you would see \\ & a machine and a molecule. \\
Reference &  \selectlanguage{russian}Дивлячись на них, ви побачите \\ & \selectlanguage{russian}машину і молекулу. \\
\textsc{Pivot} & \selectlanguage{russian}І бачити, дивлячись на них, ви \\ & \selectlanguage{russian}бачите машину і молекулу \\ & \selectlanguage{russian}молекули. \\
& (And to see, looking at them, you see \\ & a machine and a molecule of a \\ & molecule.) \\
\textsc{Softmax} & \selectlanguage{russian} І так, дивлячись на них, ви \\ & \selectlanguage{russian} бачите машину і молекулу. \\
& (And so, looking at them, you see a \\ & machine and a molecule.) \\ 
\ours &  \selectlanguage{russian} І дивляючись на них, ви побачите\\ & \selectlanguage{russian}машину і молекулу. \\
& (And looking at them, you will see a \\ & machine and a molecule) \\ 
\midrule
Source &  They have exactly the same amount of \\ & carbon. \\
Reference &  \selectlanguage{russian}Вони мають однакову \\ & \selectlanguage{russian}кількість вуглецю. \\
\textsc{Pivot} & \selectlanguage{russian}Таким чином, їх частка вуглецю.  \\
& (Thus, their share of carbon.) \\
\textsc{Softmax} & \selectlanguage{russian}Вони мають однакову кількість \\ & \selectlanguage{russian} вуглецю. \\ 
& (They have the same amount of carbon.) \\ 
\ours &  \selectlanguage{russian}Вони мають точно таку ж \\ & \selectlanguage{russian}кількість вуглецю. \\
& (they have exactly the same amount of \\& carbon) \\ 
\bottomrule
  \end{tabular}
  \caption{\small Examples of \textsc{en-uk} translations generated by \ours and the best performing baselines.}
  \label{tab:examples}
\end{table}

\begin{table}[hbt]
\small
  \centering
  \begin{tabular}{r|r}  
  \midrule
  Source & I've never heard of this address near here. \\
Reference &  \selectlanguage{arabic} \textRL{ما قط سمعت بهالعنوان في ها}\\
& \selectlanguage{arabic}\textRL{لمنطقة من قبل.} \\
\textsc{Pivot} & \selectlanguage{arabic}\textRL{ رح يسلمك.}  \\
& (He will hand over.)\\
\textsc{Softmax} & \selectlanguage{arabic} \textRL{ولا مرة سمعت عن هالعنوان هني.} \\
& (Not once did I hear this title here)\\
& \selectlanguage{arabic} \textRL{ما سمعت أبدًا من هعنوان قريب من هون.} \\
& (I've never heard from this address near \\ & here.)\\\midrule
Source & What's the exchange rate today? \\
Reference & \selectlanguage{arabic} \textRL{شنو السعر اليوم؟ } \\
\textsc{Pivot} & \selectlanguage{arabic} \textRL{سعر اليوم؟}\\
& (What's the rate?)\\
\textsc{Softmax} & \selectlanguage{arabic} \textRL{شنو سعر الصرف اليوم؟ } \\
& (What's the exchange rate today?) \\
 & \selectlanguage{arabic} \textRL{شو سعر الصرف اليوم؟}\\
& (What's the exchange rate today?)\\
\midrule
Source & How do I get to that place? \\
Reference & \selectlanguage{arabic} \textRL{كيف بوصل لهالمطرح؟}\\
\textsc{Pivot} & \selectlanguage{arabic} \textRL{كيف بتنصح؟}\\
& (How do you recommend?) \\
\textsc{Softmax} & \selectlanguage{arabic} \textRL{كيف فيي أوصل عالمحل؟}\\
& (How can I get to the shop?) \\
& \selectlanguage{arabic} \textRL{كيف فيي وصل؟}\\ 
& (How can I get there?) \\ \midrule
Source & Tell me when we get to the museum. \\
Reference & \selectlanguage{arabic} \textRL{قلي بس نوصل عالمتحف.}\\
\textsc{Pivot} & \selectlanguage{arabic} \textRL{رح نروح عالتاني}\\
& (we will go to the other.) \\
\textsc{Softmax} & \selectlanguage{arabic} \textRL{احكي ايمتى نوصل عالمتحف.}\\
& (Talk when we get to the museum) \\
& \selectlanguage{arabic} \textRL{قلي ايمتى وصلنا للمتحف.}\\
& (Tell me when we got to the museum)\\ \midrule
Source & Please take me to the morning market. \\
Reference & \selectlanguage{arabic} \textRL{عمول معروف خدني على سوق الصبح.}\\
\textsc{Pivot} & \selectlanguage{arabic} \textRL{رح نطرني.}\\
& (We'll wait) \\
\textsc{Softmax} & \selectlanguage{arabic}\textRL{ منتاخدني عالسوق الصبح.} \\
& (You take us to the market this morning.) \\
& \selectlanguage{arabic}\textRL{ منفضل تاخدني عالسوق الصبح.}
\\
& (We prefer you take us to the market at the \\ & morning.) \\ \midrule
  \end{tabular}
  \caption{\small Examples of English to Beirut Arabic translations generated by \ours and the best performing baselines.}
  \label{tab:beirut_examples}
\end{table}

\section{Negative Results}
\label{app:negative-results}

We present results for the following experiments: (a)~adapting an English to Thai (\textsc{en\into th}) model to Lao (\textsc{lo}). We use a parallel corpus of around 10M sentences for training the supervised \textsc{en\into th} model from the CCAligned corpus~\citep{elkishky_ccaligned_2020}, around 140K \textsc{lo} monolingual sentences from the OSCAR corpus~\citep{ortiz-suarez-etal-2020-monolingual} and TED2020 dev/tests for both \textsc{th} and \textsc{lo}\footnote{Although Thai and Lao scripts look very similar, they use different Unicode symbols which are one-to-one mappable to each other:~\url{https://en.wikipedia.org/wiki/Lao_(Unicode_block)}}~\citep{reimers-2020-multilingual-sentence-bert}.  (b)~adapting an English to Amharic Model (\textsc{en\into am}) to Tigrinya (\textsc{ti}). We use training, development and test sets from the JW300 corpus~\citep{agic-vulic-2019-jw300} containing 500K \textsc{en--am} parallel corpus and 100K Tigrinya monolingual sentences.

\begin{table}[hbt]
\small
\centering
\begin{tabular}{lc|c}
\toprule
& \textsc{en\into lo} & \textsc{en\into ti}\\ \toprule
\textsc{src\into std} & 0.7 & 1.8\\ 
\textsc{Softmax} & 1.4 & 2.9\\
\textsc{LangVarMT} & 4.5 & 3.8\\ \bottomrule
\end{tabular}
\caption{\small BLEU scores for English to Lao and English to Tigrinya translation}
\label{tab:negative}
\end{table}


As summarized in \Tref{tab:negative}, our method fails to perform well on these sets of languages. Although Thai and Lao are very closely related languages, we attribute this result to little subword overlap in their respective vocabularies which degrade the quality of the embeddings. This is because Lao's writing system is developed phonetically whereas Thai writing contains many silent characters. 
Considering shared phonetic information while learning the embeddings can alleviate this issue and is an avenue for future work. On the other hand, Amharic and Tigrinya, while sharing a decent amount of vocabulary, use different constructs and function words~\citep{kidane2021exploration} leading to a very noisy psuedo-parallel corpus. 






\section{Measuring Unfairness}
\label{app:unfairness}

When evaluating multilingual and multi-dialect systems, it is crucial that the evaluation takes into account principles of fairness, as outlined in economics and social choice
theory~\cite{choudhury-deshpande-2021-linguistically}. 
We follow the least difference principle proposed by~\citet{rawls}, whose egalitarian approach proposes to narrow the gap between unequal accuracies. 

A simple proxy for unfairness is the standard deviation (or, even simpler, a $\max-\min$ performance) of the scores across languages. Beyond that, we measure a system's \textit{unfairness} with respect to the different subgroups using the adaptation of generalized entropy index described by~\citet{speicher2018unified}, which considers equities within and between subgroups in evaluating the overall unfairness of an algorithm on a population. The generalized entropy index for a population of $n$ individuals receiving benefits $b_1,b_2,\ldots,b_n$ with mean benefit $\mu$ is
\begin{align*}
    \mathcal{E}^\alpha(b_1,\ldots,b_n) = \frac{1}{n\alpha(\alpha-1)}\sum_{i=1}^n \left[ \left(\frac{b_i}{\mu}\right)^\alpha -1\right].
\end{align*}
Using $\alpha=2$ following \citet{speicher2018unified}, the generalized entropy index corresponds to half the squared coefficient of variation.\footnote{The coefficient of variation is simply the ratio of the standard deviation $\sigma$ to the mean $\mu$ of a distribution.} 

If the underlying population can be split into $|G|$ disjoint subgroups across some attribute (e.g. gender, age, or language variety) we can decompose the total unfairness into individual and group-level unfairness. Each subgroup $g\in G$ will correspond to $n_g$ individuals with corresponding benefit vector $\mathbf{b}^g = (b^g_1,b^g_2,\ldots,b^g_{n_g})$ and mean benefit $\mu_g$. Then, total generalized entropy can be re-written as:
\begin{align*}
    \mathcal{E}^{\alpha}(b_1,\ldots,b_n) = &\sum_{g=1}^{|G|}\frac{n_g}{n}\left( \frac{\mu_g}{\mu}\right)^{\alpha} \mathcal{E}^{\alpha}(\mathbf{b}^g) \\
    + \sum_{g=1}^{|G|}\frac{n_g}{n\alpha(\alpha-1)}&\left[ \left(\frac{\mu_g}{\mu}\right)^\alpha -1\right]\\
    =& \mathcal{E}^{\alpha}_{\omega}(\mathbf{b}) + \mathcal{E}^{\alpha}_{\beta}(\mathbf{b}).
\end{align*}
The first term $\mathcal{E}^{\alpha}_{\omega}(\mathbf{b})$ corresponds to the weighted unfairness score that is observed \textit{within} each subgroup, while the second term $\mathcal{E}^\alpha_\beta(\mathbf{b})$ corresponds to the unfairness score \textit{across} different subgroups. 

In this measure of unfairness, we define the benefit as being directly proportional to the system's accuracy. For a Machine Translation system, each user receives an average benefit equal to the BLEU score the MT system achieves on the user's dialect.
Conceptually, if the system produces a perfect translation (BLEU=1) then the user will receive the highest benefit of 1. If the system fails to produce a meaningful translation (BLEU$\rightarrow0$) then the user receives no benefit ($b=0$) from the interaction with the system.

\end{document}